\documentclass[conference]{IEEEtran}
\usepackage{cite}
\usepackage{amsmath,amssymb,amsfonts}
\usepackage{multicol,caption}
\usepackage{commath}
\usepackage{algorithmic}
\usepackage[ruled]{algorithm2e}
\usepackage{graphicx}
\usepackage{textcomp}
\usepackage{multirow}
\usepackage{balance}
\usepackage{xcolor}
\def\BibTeX{{\rm B\kern-.05em{\sc i\kern-.025em b}\kern-.08em
    T\kern-.1667em\lower.7ex\hbox{E}\kern-.125emX}}

\title{Anomaly Unveiled: Securing Image Classification against Adversarial Patch Attacks}
\author{\IEEEauthorblockN{Nandish Chattopadhyay, Amira Guesmi, and Muhammad Shafique}
 \IEEEauthorblockA{\textit{eBrain Lab, Division of Engineering, New York University (NYU) Abu Dhabi, UAE}} \
 } 
\begin{document}
\maketitle
\begin{abstract}
Adversarial patch attacks pose a significant threat to the practical deployment of deep learning systems. However, existing research primarily focuses on image pre-processing defenses, which often result in reduced classification accuracy for clean images and fail to effectively counter physically feasible attacks. In this paper, we investigate the behavior of adversarial patches as anomalies within the distribution of image information and leverage this insight to develop a robust defense strategy. Our proposed defense mechanism utilizes a clustering-based technique called DBSCAN to isolate anomalous image segments, which is carried out by a three-stage pipeline consisting of Segmenting, Isolating, and Blocking phases to identify and mitigate adversarial noise. Upon identifying adversarial components, we neutralize them by replacing them with the mean pixel value, surpassing alternative replacement options. Our model-agnostic defense mechanism is evaluated across multiple models and datasets, demonstrating its effectiveness in countering various adversarial patch attacks in image classification tasks. %Comparative analysis against existing literature confirms the superiority of our approach.
Our proposed approach significantly improves accuracy, increasing from 38.8\% without the defense to 67.1\% with the defense against LaVAN and GoogleAp attacks, surpassing prominent state-of-the-art methods such as LGS \cite{naseer2019local} (53.86\%) and Jujutsu \cite{Jujutsu} (60\%).
\end{abstract}   
\section{Introduction}
\label{intro}
%Adversarial attacks pose a formidable threat to the robustness and performance of highly trained deep neural network (DNN) models \cite{CW, LiV15, Goodfellow2015ExplainingAH}. In these attacks, an adversary strategically introduces adversarial perturbations to test samples, causing significant disruptions to the model's accurate predictions. A particularly potent manifestation of adversarial attacks involves the insertion of localized patches into test images. By exploiting this vulnerability, attackers force the DNN model to make errors in critical tasks such as image classification or object detection \cite{10268441, guesmi2023dap, Hu21}.
Adversarial manipulations pose a significant challenge to the resilience and effectiveness of well-trained deep neural network (DNN) architectures \cite{guesmi2023physical, CW, LiV15, Goodfellow2015ExplainingAH, cod_1, cod_2}. In such scenarios, adversaries strategically introduce perturbations to test samples, leading to noticeable disruptions in the model's ability to accurately predict outcomes. A notable form of these attacks involves the insertion of localized patches into test images, exploiting vulnerabilities and causing the DNN model to err in crucial tasks such as image classification or object detection \cite{ guesmi2023dap, guesmi2023advart, Hu21}.

%Patch-based attacks stand out as a widely recognized and practical form of adversarial attacks due to their adaptability, particularly in situations with limited access \cite{10268441}. In contrast to conventional adversarial techniques that require extensive perturbations across the entire target object, patch-based attacks showcase a localized nature. These attacks operate like discreet stickers, allowing them to be effortlessly applied to potential targets—simulating real-world scenarios where adversaries may encounter constraints in resources or access. The subtle characteristics of patch-based attacks contribute to their elusive nature, emphasizing the urgent need for the swift implementation of robust defense mechanisms. 
%However, these defenses are susceptible to generating false positives \cite{naseer2019local} and exhibit difficulty in accurately distinguishing between adversarial examples and clean samples. Additionally, in some cases, these defenses might remove or alter important features \cite{xiang2021patchguard, levine2020randomized}, causing the model's performance to degrade even on benign samples.
Patch-based attacks are recognized as a practical form of adversarial manipulation, valued for their adaptability, especially in scenarios with limited accessibility \cite{guesmi2023physical, guesmi2024saam, guesmi2023aparate}. Unlike traditional adversarial methods that require extensive perturbations spanning the entire target object, patch-based attacks exhibit a localized nature. These attacks function like discrete stickers, making them easy to apply to potential targets, reflecting real-world situations where adversaries may face resource or access constraints. The subtle attributes of patch-based attacks contribute to their elusive nature, emphasizing the urgent need for the rapid deployment of robust defense mechanisms.

However, these defenses are prone to generating false positives \cite{naseer2019local} and face challenges in accurately distinguishing between adversarial and clean samples. Additionally, in certain instances, these defenses may inadvertently remove or alter crucial features \cite{xiang2021patchguard, levine2020randomized}, resulting in the degradation of the model's performance even on benign samples.

Adversarial patches exhibit characteristics of outliers or anomalies within the distribution of input images. The adversarial noise embedded within these patches diverges significantly from the signal or information present in the rest of the sample. Leveraging advanced anomaly detection techniques facilitates the identification and segregation of these patches in instances where they deviate from the broader image distribution. This is particularly useful in developing practical adversarial defenses against such patch based attacks.

\subsection{Contribution}
The primary contributions of this paper is mentioned here:
\begin{itemize}
    \item %We have proposed a defense mechanism against adversarial patch attacks that works by isolating the region in the image containing the patch as an anomaly to the rest of the image, and blocks the adversarial information thereafter.
    We introduce a novel defense mechanism against adversarial patch attacks. Our approach involves isolating the region in the image containing the patch as an anomaly and subsequently blocking the adversarial information.
    \item We demonstrate the distinctive informational disparities inherent in adversarial patches, offering invaluable insights crucial for the development of resilient defense strategies against adversarial patch attacks.
    % the distinct informational variability 
    \item %We have used a three-step pipeline for its implementation, involving a Segmenting phase, an Isolating phase and a Blocking phase. The Segmenting phase chops the image in parts, which are used as input to a clustering algorithm (DBSCAN), which is able to identify the segments that contain the adversarial noise. We then replace the said segments with the mean pixel value to render the adversarial patch ineffective. 
    We propose a three-step pipeline for implementing our defense mechanism, comprising a Segmenting phase, an Isolating phase, and a Blocking phase. Initially, the Segmenting phase divides the image into parts, which are then subjected to a clustering algorithm (DBSCAN) \cite{dbscan, dbscan_2} to identify segments containing adversarial noise. Subsequently, we replace these identified segments with the mean pixel value to neutralize the adversarial patch.
    \item %The proposed defense mechanism is model agnostic and has shown impressive performances of up to X \% recovery on adversarial samples on image classification tasks across various datasets, adversarial patches and neural architectures. 
    Our defense mechanism is model-agnostic and demonstrates impressive performance, achieving up to 85\% recovery on adversarial samples in image classification tasks across various datasets, adversarial patches, and neural architectures.

\end{itemize}

\begin{figure*}[ht!]
\centerline{\includegraphics[width=2\columnwidth]{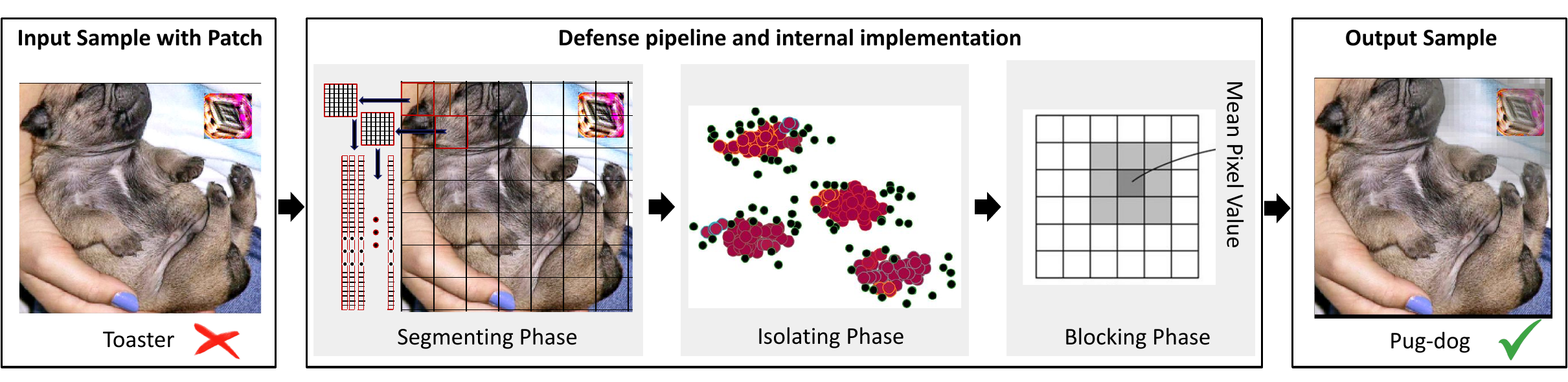}}
\caption{Detailed diagram of our proposed methodology.}
\label{fig:methodology}
\end{figure*}

\section{Theoretical Background}
\label{theory}

\subsection{Adversarial Patches}
%Adversarial patches represent a specialized category of adversarial perturbations aimed at manipulating localized patches or specific regions within an image to deceive classification models. These attacks exploit the inherent vulnerability of models to localized alterations, with the ultimate goal of introducing subtle modifications that exert a substantial impact on the model's output. Leveraging the model's dependence on particular features or patterns, adversaries can craft patches designed to mislead the model into either misclassifying the image or perceiving it in a manner contrary to its intended interpretation.

Adversarial patches constitute a distinct subset of adversarial perturbations targeting localized areas within images to deceive classification models. These attacks exploit the susceptibility of models to localized changes, aiming to introduce subtle modifications that significantly affect the model's predictions. By exploiting the model's reliance on specific features or patterns, adversaries create patches intended to mislead the model into either misclassifying the image or interpreting it differently from its true representation.

\subsubsection{LaVAN \cite{lavan}}
%LaVAN is a technique for generating localized and visible patches that can be applied across various images and locations. This approach involves training the patch iteratively by selecting a random image and placing it at a randomly chosen location. This iterative process makes sure that the model can capture the distinguishing features of the patch across a range of scenarios, thereby enhancing its ability to transfer and its overall effectiveness.
LaVAN is a method designed to create localized and discernible patches that are applicable to a variety of images and positions. This technique entails iteratively training the patch by selecting a random image and positioning it at a randomly selected location. This iterative training ensures that the model can learn the distinctive characteristics of the patch across diverse scenarios, thereby improving its transferability and overall efficacy.

\subsubsection{GoogelAp \cite{googleap}}
%GoogelAp offers a more practical form of attack for real-world scenarios compared to Lp-norm-based adversarial perturbations, which require object capture through a camera. This attack creates universal patches that can be applied anywhere. Additionally, the attack incorporates Expectation over Transformation (EOT) \cite{eot} to enhance the strength of the generated adversarial patch.
The GoogleAp technique offers a pragmatic method for real-world attacks, contrasting with Lp-norm-based adversarial perturbations, which require object capture using a camera. GoogleAp generates universal patches adaptable to diverse locations. Additionally, the attack incorporates Expectation over Transformation (EOT) \cite{eot} to enhance the effectiveness of the generated adversarial patch.

\subsection{Attack Formulation}
%In the context of image classification, consider a deep learning-based image classifier represented as $f: X \rightarrow Y$, which is the mapping of an input image $x$ from the set of images $X$ to an output class with label $y$ from the set of labels $Y$. An adversarial example, denoted as $x^*$, is given by:
In the context of image classification, let's consider a deep learning-based image classifier denoted as $f: X \rightarrow Y$. Here, $X$ represents the set of images, and $Y$ represents the set of labels. The classifier maps an input image $x$ from $X$ to an output class with label $y$ from $Y$. An adversarial example, represented as $x^*$, is defined as:
\begin{equation}
\label{eq:adv}
     \begin{array}{lll}
         x^* \in X, \quad f(x) = y, \quad f(x^*) = y^*, \quad y \neq y^*  \nonumber 
     \end{array}
\end{equation}

%Here, $y^*$ is the targeted label, and $x^*$ is the adversarial example generated from the original input $x$. In the context of patch-based attacks, a portion of the image is replaced by the patch denoted as $P$.
Here, $y^*$ represents the targeted label, and $x^*$ is the adversarial example generated from the original input $x$. Specifically in the context of patch-based attacks, a segment of the image is substituted with the patch, symbolized as $P$.

Technically, the formulation of an adversarial example with a generated patch is expressed as:

\begin{equation}
    x^* = (1 - m_P) \odot x + m_P \odot P  \nonumber 
\end{equation}

%Here, $\odot$ represents component-wise multiplication, $P$ is the adversarial patch, and $m_P$ is a mask matrix that constrains the shape, size, and pasting position of the patch. The value of the pasting area is set to 1, and 0 elsewhere.
Here, $\odot$ denotes component-wise multiplication, where $P$ represents the adversarial patch, and $m_P$ is a mask matrix that governs the shape, size, and pasting position of the patch. Specifically, the pasting area is assigned a value of 1, while the remaining areas are set to 0.
%To ensure that the patch $P$ is input-agnostic, it is trained over a variety of images. In the LaVAN approach \cite{lavan}, the patch is trained for a fixed location for each input $x \in X$. In the case of GoogleAp, the patch is trained to be applied in any random location. To further enhance the robustness of patch $P$ and make it physically realizable, GoogleAp \cite{googleap} uses a EOT framework \cite{eot}. EOT or Expectation over Transformation essentially uses various environmental transforms $T$ that can alter $x$ in various physical environments, such as translation, rotation, or lightness changes. Adversarial examples generated under these different transformations aim to remain robust, thus enhancing the overall effectiveness of the attack.
To ensure the patch $P$ remains input-agnostic, it undergoes training across a range of images. LaVAN \cite{lavan} adopts a fixed location strategy, where the patch is trained to stay at a predetermined position for each input $x \in X$. Conversely, GoogleAp employs a random location approach, training the patch to be applied at any position within the image. Furthermore, to enhance the robustness and physical realizability of patch $P$, GoogleAp utilizes an Expectation over Transformation (EOT) framework \cite{eot}. EOT encompasses various environmental transformations $T$, such as translation, rotation, or changes in lighting conditions, applied to $x$. Adversarial examples generated under these diverse transformations aim to maintain their efficacy, thereby enhancing the attack's overall effectiveness.
% \subsubsection{LaVAN \cite{lavan}}
% LaVAN is a technique for generating localized and visible patches that can be applied across various images and locations. This approach involves training the patch iteratively by selecting a random image and placing it at a randomly chosen location. This iterative process makes sure that the model can capture the distinguishing features of the patch across a range of scenarios, thereby enhancing its ability to transfer and its overall effectiveness.

% \subsubsection{GoogelAp \cite{googleap}}
% GoogelAp offers a more practical form of attack for real-world scenarios compared to Lp-norm-based adversarial perturbations, which require object capture through a camera. This attack creates universal patches that can be applied anywhere. Additionally, the attack incorporates Expectation over Transformation (EOT) \cite{eot} to enhance the strength of the generated adversarial patch.

\subsection{Adversarial patches as anomalies}
\begin{figure}[htbp]
\centerline{\includegraphics[width=\columnwidth]{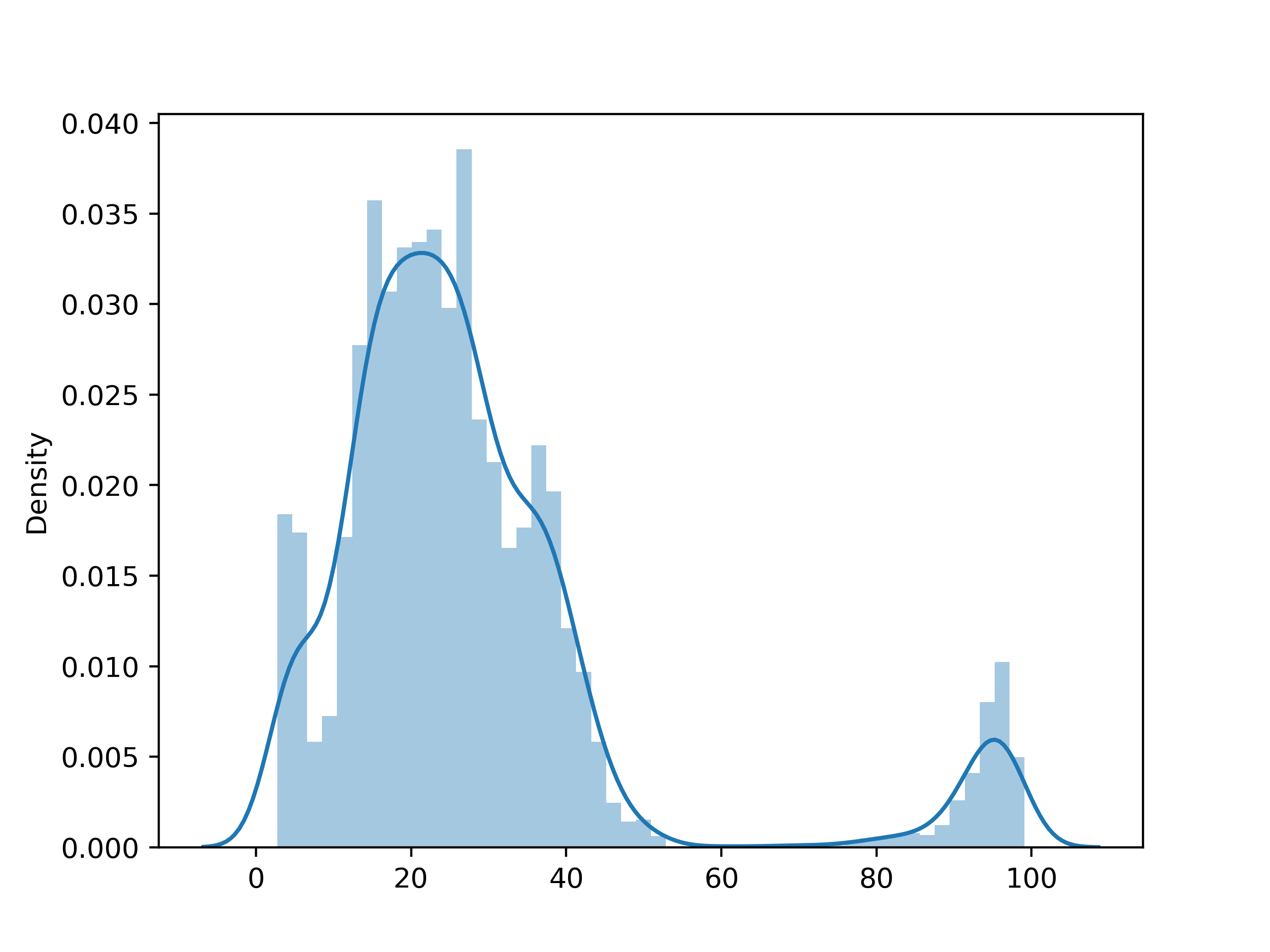}}
\caption{Plot of Mahalanobis distances of segments to represent anomalous behaviour of adversarial patches as seen in the bi-model distribution.}
\label{fig:mahal}
\end{figure}

To study whether the adversarial noise contained in the adversarial patches belong to a different distribution or not, as compared to the clean images, we make use of a distance metric to measure the distance between the overall distribution of the image and the adversarial patch. Specifically, we split the adversarial image along with the patch into segments (as described in Section \ref{method}) and fit a distribution to it, and calculate the Mahalanobis distance of every segment from that distribution \cite{mahal_1}. 
The Mahalanobis distance \cite{mahal} is a distance metric which is designed to measure the distances of data points with respect to a distribution. Formally, for a probability distribution $Q$ on $\mathbb{R}$, which has the mean $\mu = (\mu_{1},\ldots, \mu_{N})^T$, a positive definite covariance matrix $S$, then for any point $x = (x_{1}, \ldots, x_{N})^T$ for the said distribution $Q$, the Mahalanobis distance is \cite{mahal_2}: 
\begin{equation}
    d_{M} (x,Q) = \sqrt{(x-\mu)^T S^{-1}(x-\mu)} \nonumber
\end{equation}
For our case, the $x_{i}$s are the segments created as part of the Segmentation phase. Once we calculate the Mahalanobis distance for every such segment, and plot them, we observe a bi-modal distribution in Figure \ref{fig:mahal}, with the distance measured for the segments lying on the adversarial patch having a significantly higher value of Mahalanobis distance than the rest. This is a clear indication that the patch contains informational variability that is different from the rest of the image and can be isolated as anomalies.

\subsection{Anomaly Detection with DBSCAN}
We make use of the insight that adversarial patches contain anomalous features when compared to clean image samples to propose a defense mechanism. This is achieved using an unsupervised clustering technique called DBSCAN \cite{dbscan}. This is a density based algorithm for discovering clusters in large spatial datasets with noise.  
The DBSCAN algorithm \cite{dbscan_2} takes a dataset, denoted as $data$, a distance function \textit{distFunc} and two hyper-parameters $eps$ and $minPts$ as inputs. The $eps$-neighbourhood of any data sample $p$ is denoted by $N_{eps}(p)$, and is given by the following condition: $N_{eps}(p)$ = $\{q$ $\in$ $data$ $|$ $distFunc(p,q)$ $\leq eps \}$. Also, this data point $p$ in consideration is \textit{directly density reachable} as per definition if both the conditions $p \in N_{eps}(q)$ and $|N_{eps}(q)| \geq minPts$ are simultaneously satisfied. Any data point $p$ is \textit{density reachable} if there are points $p_{1}, \ldots, p_{n}$, $p_{1} \approx q$, $p_{n} = p$, such that $p_{i+1}$ is directly \textit{density reachable} from $p_{i}$. Similarly, the data point $p$ is \textit{density connected} to a point $q$, if there exists a point $o$ such that both $p$ and $q$ are \textit{density reachable} from $o$, for the set $eps$ and $minPts$ value. These definitions help us in formalizing the building of the clusters using the DBSCAN algorithm. A cluster $C$ (where $C \subseteq data$) is formed from the $data$ if the following conditions are satisfied: $\forall $ $p,q$, if $p \in C$ and $q$ is \textit{density reachable} from $p$ for the corresponding $eps$ and $minPts$, then $q \in C$; and $\forall$ $p,q \in C$, $p$ is \textit{density connected} to $q$ for the corresponding $eps$ and $minPts$. These two conditions are called the Maximality and Connectivity conditions. If we consider the $k$ clusters so generated to be $C_{1}, \ldots, C_{k}$, then the anomalies are the data points that have the label \textit{Noise}, where those points do not belong to any cluster $C_{i}$, such that $Noise$ = $\{p \in data$ $|$ $\forall i $ where $p \notin C_{i}$ $\}$ for a set $eps$ and $minPts$ value \cite{dbscan_3,dbscan_4}. 
The details of the algorithm is mentioned in Algo \ref{dbscan}.

\begin{algorithm}
\SetAlgoLined
\caption{$DBSCAN(data, distFunc, eps, minPts$)}
\label{dbscan}
$C \gets 0$ \\
$\bold{for}$ $(P \in  data)$: \\
\Indp $\bold{if}$ (label($P$) $\neq$ undefined): \\
\Indp $N$ $\gets$ Search$(data, distFunc, P, eps)$ \\
\Indm 
$\bold{end}$ $\bold{if}$ \\
$\bold{if}$ ($|N| < minPts)$: \\
\Indp label ($P$) $\gets$ Noise \\
\Indm 
$\bold{end}$ $\bold{if}$ \\
$C \gets C + 1$ \\
label($P$) $\gets$ $C$ \\
Set $S$ $\gets$ $N \backslash \{P\}$ \\
$\bold{for}$ $(Q \in  S)$: \\
\Indp $\bold{if}$ (label($Q$) == Noise): \\
\Indp label($Q$) $\gets$ $C$ \\
\Indm
$\bold{end}$ $\bold{if}$ \\
$\bold{if}$ (label($Q$) == undefined): \\
\Indp label($Q$) $\gets$ $C$ \\
$N$ $\gets$ Search($data, distFunc, Q, eps$) \\
$\bold{if}$ ($|N|$ $\geq$ $minPts$): \\
\Indp $S \gets S \bigcup N $ \\
\Indm
$\bold{end}$ $\bold{if}$ \\
\Indm
$\bold{end}$ $\bold{if}$ \\
\Indm
$\bold{end}$ $\bold{for}$ \\
\Indm 
$\bold{end}$ $\bold{for}$ \\
Initialize $Y = (y_{1}, \ldots, y_{n})$ \\
$\bold{for}$ $(i \in  (1,\ldots,n))$: \\
\Indp $y_{i}$ $\gets$ label($P$) \\
\Indm
$\bold{end}$ $\bold{for}$ \\
$\bold{return}$ $Y$
\end{algorithm}

\begin{algorithm}
\SetAlgoLined
\caption{$Search(data, distFunc, Q, eps$)}
\label{search}
$N \gets Q$ \\
$\bold{for}$ $(P \in  data)$: \\
\Indp $\bold{if}$ ($distFunc$($Q,P$)) $\leq$ $eps$): \\
\Indp $N \gets N \bigcup \{P\}$ \\
\Indm
$\bold{end}$ $\bold{if}$ \\
\Indm
$\bold{end}$ $\bold{for}$ \\
$\bold{return}$ $N$
\end{algorithm}
\section{Defense Technique}
\label{method}

As illustrated in Figure \ref{fig:methodology}, our proposed defense mechanism has a three stage pipeline, that we have created to isolate the area of the image containing the adversarial patch and thereby, mitigating it. The first part of the process is the Segmenting Phase, which chops up the image into kernels using a moving window, which parses the entire image. Thereafter, in the Isolating Phase, the generated segments are used as inputs to the clustering mechanism. The segments identified as the anomalies are subjected to the Blocking Phase, where the adversarial noise noise is destroyed. The complete algorithm is described in details in Algo \ref{algo}.  

\subsection{Segmenting Phase}
We parse the image using a moving window of a fixed kernel size and a specific stride length, and chop us the image into segments which have partial overlap. Since the kernel size and the stride length are both hyper-parameters to the system, the amount of overlap can be varied. These segments are converted to long vectors, which is used as the data samples for the clustering algorithm subsequently. 

\subsection{Isolating Phase}
The segments, generated through the earlier step, serve as the data samples on which the anomaly detection method is applied. It may be noted here that the goal is to identify those segments which correspond to the part of the image that contains the adversarial patch. We propose to use clustering for this purpose, specifically DBSCAN \cite{dbscan}, which is very efficient in identifying noise or anomalous samples, which do not belong to any of the clusters. Notably, for the most optimal performance, we tune the parameters of the DBSCAN algorithm, which are $eps$ and the minimum number of points in a cluster $minPts$. The details and the significance of these hyper-parameters and how they help in the DBSCAN clustering is explained in details in Section \ref{theory}. Once we run the DBSCAN algorithm, the labels corresponding to all the samples are noted (labels signify the associated cluster ID for each sample/segment) and the samples which have been labelled as \textit{Noise} are identified as the anomalies. 

\subsection{Blocking Phase}
Once some of the segments are identified as the those with anomalous behaviour with respect to the other set of segments corresponding to the rest of the image, we use the \textit{Replace}() function to set all pixels within the segment to a common value. We experimented with the minimum, mean and maximum values of the segment and our observations provide evidence of superior performance of the mean pixel values. Therefore, in all our experiments, we have used the average value of the pixels (channel-wise for the colours). 

\begin{algorithm}
\SetAlgoLined
\caption{Anomaly Detection and Mitigation}
\label{algo}
$\boldsymbol{IN}:$ $S$: sample image, $k$: size of kernel, $strlen$:  length of stride, $eps$: DBSCAN parameter, $minPts$: minimum size of clusters\\
$\boldsymbol{OUT}:$ $S'$: Image with neutralised patch \\ 
/*$\boldsymbol{Segmenting\_Phase}$*/\\
Create $n$ segments $ X \gets (x_1, \ldots, x_n)$ from sample $S$, using  = $k$ as kernel size and $strlen$ as stride length \\ 
/*$\boldsymbol{Isolating\_Phase}$*/\\
$Y = DBSCAN(X,eps,minPts)$ (Algo \ref{dbscan}) \\
$\boldsymbol{initialize}$ $ A = \{\}$ \\
$\bold{for}$ $(i \in (1,\ldots,n)$: \\
\Indp
$\bold{if}$ ($y_{i} == Noise$): \\
\Indp $A \gets x_{i}$ \\
\Indm 
$\bold{end}$ $\bold{if}$ \\
\Indm 
$\bold{end}$ $\bold{for}$ \\ 
/*$\boldsymbol{Blocking\_Phase}$*/\\
Set $r = Size(A)$ \\
$\bold{for}$ $(j \in (1,\ldots,r))$: \\
\Indp $x_{j}$ $\gets$ $Replace(x_{j})$ \\
\Indm 
$\bold{end}$ $\bold{for}$ \\
Replace $r$ segments into $S$ to generate $S'$ \\
$\boldsymbol{return}$ $S'$
\end{algorithm}
\section{Evaluation}
\label{results}
%In this section, we thoroughly assess the effectiveness of our proposed defense mechanism. 
In this section, we conduct a comprehensive evaluation of the efficacy of our proposed defense mechanism.

\subsection{Experimental Setup}
\subsubsection{Datasets and Networks}
%We conducted our defense evaluation on ImageNet \cite{imagenet} using three pretrained deep neural networks (DNNs) available in the TorchVision library: Resnet-50 \cite{he2015deep}, Resnet-152 \cite{he2015deep}, and VGG-19 \cite{simonyan2015deep}. These well-established models served as the basis for our assessment, ensuring a comprehensive evaluation across a range of DNN architectures and model complexities.
We conducted our defense evaluation on the ImageNet dataset \cite{imagenet}, employing three pretrained deep neural networks (DNNs) accessible in the TorchVision library: ResNet-50 \cite{he2015deep}, ResNet-152 \cite{he2015deep}, and VGG-19 \cite{simonyan2015deep}. These widely recognized models were selected to facilitate a thorough assessment across various DNN architectures and model complexities.
\subsubsection{Attack Setup}
\begin{figure}[htbp]
\centerline{\includegraphics[width=0.6\columnwidth]{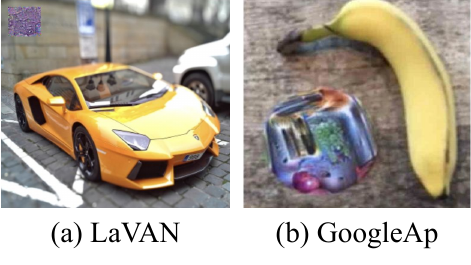}}
\caption{Patch based attacks: a) LaVAN \cite{lavan}, b) GoogleAp \cite{googleap}.}
\label{fig:attacks}
\end{figure}

%The attacker's objective is to create adversarial patches that effectively deceive the victim deep learning-based classifier. We generate distinct patches in five different sizes: $38 \times 38$, $41 \times 41$, $44 \times 44$, $47 \times 47$, and $50 \times 50$. In the case of the LaVAN patch, the patch's location is fixed to the upper right corner of the image. On the other hand, for GoogleAp, the patch is placed randomly within the image. Each patch undergoes a training process comprising 100 epochs using a training dataset consisting of 1000 images. Subsequently, we assess the attack success rate on a separate test dataset. In the context of ImageNet, our evaluation employs a set of 10,000 images drawn from the validation dataset.
The attacker aims to create adversarial patches capable of effectively deceiving the victim deep learning-based classifier. We generate distinct patches in five different sizes: $38 \times 38$, $41 \times 41$, $44 \times 44$, $47 \times 47$, and $50 \times 50$. In the case of the LaVAN patch, the patch's location is fixed to the upper left corner of the image. Conversely, for GoogleAp, the patch is randomly placed within the image (refer to Figure \ref{fig:attacks}). Each patch undergoes a training process consisting of 100 epochs, utilizing a training dataset comprising 1000 images. Subsequently, we evaluate the attack success rate on a separate test dataset. In the context of ImageNet, our evaluation encompasses a set of 10,000 images sampled from the validation dataset.
\subsubsection{Defense Setup}
%The defense mechanism comprises of two phases, as shown in Section \ref{method}. In the Training phase, we use a $50-50\%$ proportion of samples of $1000$, from the ImageNet validation dataset to fine tune the model and also set the optimal working parameter $I$ for the dimension reduction block. We systematically experimented with different values of the Variability parameter, specifically considering settings $I \in 99\%, \ldots, 90\%$. Through empirical analysis, we identified the parameter value that achieves the optimal balance between robust accuracy, which measures the model's performance under adversarial conditions, and baseline accuracy, which reflects its performance on clean, unaltered data. In the Inference phase, the samples are passed through the dimension reduction block, followed by the actual model for classification. We have repeated this setup for the combination of models and attacks mentioned below.

The defense mechanism comprises of three phases, as described in Section \ref{method}. For the Segmenting Phase, we varied the kernel size from $40 \times 40$ pixels to $20 \times 20$ pixels and a stride length $5$ to $10$ pixels and the most optimal performance with respect to downstream accuracy on the task was observed for kernel size $40 \times 40$ and stride length $8$ pixels. For the Isolating Phase, for the DBSCAN algorithm, we tuned the hyper-parameters with respect to downstream accuracy and used $eps = 0.4$ and $minPts = 1201$. 

\subsection{Experimental Results}
In our evaluation, we primarily focused on quantifying the model's robust accuracy. To demonstrate the efficacy of our defense strategy, we initially generated adversarial patches using two distinct attack methodologies: LAVAN and GoogleAp. Subsequently, we assessed the model's robust accuracy across various patch sizes and different models. The baseline performances of our defense on clean samples without patches were measured as follows: $81.2\%$ (ResNet-152), $78.4\%$ (ResNet-50), and $74.2\%$ (VGG-19) on the ImageNet dataset, and $94.1\%$ (ResNet-152), $90.9\%$ (ResNet-50), and $88.6\%$ (VGG-19) on the CalTech-101 dataset.

Tables \ref{tab1_imagenet_sup} and \ref{tab2_imagenet_sup} demonstrate the notable effectiveness of our defense technique, illustrating a remarkable level of robust accuracy in mitigating GoogleAp and LAVAN attacks on the ImageNet dataset. For instance, our defense achieves robust accuracy rates of $67.1\%$ and $63.6\%$ when employed against GoogleAp and LAVAN attacks, respectively, for the ResNet-50 model and a patch size $38 \times 38$.

\begin{table}[!htbp]
\caption{Robustness on GoogleAp attack (ImageNet dataset)}
\label{tab1_imagenet_sup}

\resizebox{0.45\textwidth}{!}{%
\begin{tabular}{c|l|c|c|c}
\hline
\multirow{2}{*}{\begin{tabular}[c]{@{}c@{}}Patch\\ Size\end{tabular}} & \multicolumn{1}{c|}{\multirow{2}{*}{\begin{tabular}[c]{@{}c@{}}Model /\\ Neural Network\end{tabular}}} & \multirow{2}{*}{\begin{tabular}[c]{@{}c@{}}Baseline\\ Accuracy\end{tabular}} & \multirow{2}{*}{\begin{tabular}[c]{@{}c@{}}Adversarial\\ Accuracy\end{tabular}} & \multirow{2}{*}{\begin{tabular}[c]{@{}c@{}}Robustness\\ (w/ patch)\end{tabular}} \\
 & \multicolumn{1}{c|}{} &  &  &  \\ \hline %\hline 
\multirow{3}{*}{\begin{tabular}[c]{@{}c@{}}38\\ x\\ 38\end{tabular}} & ResNet 152 & 81.2\% & 39.9\%  & 64.8\%  \\ %\cline{2-5} 
 & ResNet 50 & 78.4\%  & 38.8\%  & 67.1\%  \\ %\cline{2-5} 
 & VGG 19 & 74.2\%  & 39.1\%  & 63.7\%  \\ \hline
\multirow{3}{*}{\begin{tabular}[c]{@{}c@{}}41\\ x\\ 41\end{tabular}} & ResNet 152 & 81.2\% & 21.4\% & 67.1\% \\ %\cline{2-5} 
 & ResNet 50 & 78.4\% & 21.1\% & 67.2\% \\ %\cline{2-5} 
 & VGG 19 & 74.2\% & 22.8\% & 61.4\% \\ \hline
\multirow{3}{*}{\begin{tabular}[c]{@{}c@{}}44\\ x\\ 44\end{tabular}} & ResNet 152 & 81.2\%  & 14.6\%  & 70.8\%  \\ %\cline{2-5} 
 & ResNet 50 & 78.4\%  & 14.2\%  & 60.9\%  \\ %\cline{2-5} 
 & VGG 19 & 74.2\%  & 15.8\%  & 63.9\%  \\ \hline
\multirow{3}{*}{\begin{tabular}[c]{@{}c@{}}47\\ x\\ 47\end{tabular}} & ResNet 152 & 81.2\% & 9.3\% & 69.4\% \\ %\cline{2-5} 
 & ResNet 50 & 78.4\% & 9.0\% & 64.7\% \\ %\cline{2-5} 
 & VGG 19 & 74.2\% & 10.6\% & 60.3\% \\ \hline
\multirow{3}{*}{\begin{tabular}[c]{@{}c@{}}50\\ x\\ 50\end{tabular}} & ResNet 152 & 81.2\%  & 4.9\%  & 65.8\%  \\ %\cline{2-5} 
 & ResNet 50 & 78.4\%  & 4.5\%  & 67.4\%  \\ %\cline{2-5} 
 & VGG 19 & 74.2\%  & 3.8\%  & 63.1\%  \\ \hline
\end{tabular}
}
\end{table}

\begin{table}[!htbp]
\caption{Robustness on LAVAN attack (ImageNet dataset)}
\label{tab2_imagenet_sup}
\resizebox{0.45\textwidth}{!}{%
\begin{tabular}{c|l|c|c|c}
\hline
\multirow{2}{*}{\begin{tabular}[c]{@{}c@{}}Patch\\ Size\end{tabular}} & \multicolumn{1}{c|}{\multirow{2}{*}{\begin{tabular}[c]{@{}c@{}}Model /\\ Neural Network\end{tabular}}} & \multirow{2}{*}{\begin{tabular}[c]{@{}c@{}}Baseline\\ Accuracy\end{tabular}} & \multirow{2}{*}{\begin{tabular}[c]{@{}c@{}}Adversarial\\ Accuracy\end{tabular}} & \multirow{2}{*}{\begin{tabular}[c]{@{}c@{}}Robustness\\ (w/ patch)\end{tabular}} \\
 & \multicolumn{1}{c|}{} &  &  &  \\ \hline %\hline 
\multirow{3}{*}{\begin{tabular}[c]{@{}c@{}}38\\ x\\ 38\end{tabular}} & ResNet 152 & 81.2\%  & 10.1\%  & 63.1\%  \\ %\cline{2-5} 
 & ResNet 50 & 78.4\%  & 10.2\%  & 63.6\%  \\ %\cline{2-5} 
 & VGG 19 & 74.2\%  & 11.1\%  & 60.5\%  \\ \hline
\multirow{3}{*}{\begin{tabular}[c]{@{}c@{}}41\\ x\\ 41\end{tabular}} & ResNet 152 & 81.2\% & 7.9\% & 68.2\% \\ %\cline{2-5} 
 & ResNet 50 & 78.4\% & 8.3\% & 61.8\% \\ %\cline{2-5} 
 & VGG 19 & 74.2\% & 8.1\% & 57.1\% \\ \hline
\multirow{3}{*}{\begin{tabular}[c]{@{}c@{}}44\\ x\\ 44\end{tabular}} & ResNet 152 & 81.2\%  & 4.9\%  & 66.1\%  \\ %\cline{2-5} 
 & ResNet 50 & 78.4\%  & 4.8\%  & 64.7\%  \\ %\cline{2-5} 
 & VGG 19 & 74.2\%  & 4.8\%  & 61.8\%  \\ \hline
\multirow{3}{*}{\begin{tabular}[c]{@{}c@{}}47\\ x\\ 47\end{tabular}} & ResNet 152 & 81.2\% & 1.2\% & 65.8\% \\ %\cline{2-5} 
 & ResNet 50 & 78.4\% & 1.0\% & 68.9\% \\ %\cline{2-5} 
 & VGG 19 & 74.2\% & 1.7\% & 58.6\% \\ \hline
\multirow{3}{*}{\begin{tabular}[c]{@{}c@{}}50\\ x\\ 50\end{tabular}} & ResNet 152 & 81.2\%  & 1.9\%  & 63.2\%  \\ %\cline{2-5} 
 & ResNet 50 & 78.4\%  & 2.0\%  & 62.1\%  \\ %\cline{2-5} 
 & VGG 19 & 74.2\%  & 2.1\%  & 61.5\%  \\ \hline
\end{tabular}
}
\end{table}
Similarly, on the Caltech-101 dataset (refer to Tables \ref{tab1_caltech_sup} and \ref{tab2_caltech_sup}), our defense exhibits outstanding performance, achieving robust accuracy rates of $76.5\%$ and $73.6\%$ when countering GoogleAp and LAVAN attacks, respectively, for the ResNet-50 model with a patch size of $38 \times 38$.

\begin{table}[!htbp]
\caption{Robustness on GoogleAp attack (CalTech-101)}
\label{tab1_caltech_sup}
\resizebox{0.45\textwidth}{!}{%
\begin{tabular}{c|l|c|c|c}
\hline
\multirow{2}{*}{\begin{tabular}[c]{@{}c@{}}Patch\\ Size\end{tabular}} & \multicolumn{1}{c|}{\multirow{2}{*}{\begin{tabular}[c]{@{}c@{}}Model /\\ Neural Network\end{tabular}}} & \multirow{2}{*}{\begin{tabular}[c]{@{}c@{}}Baseline\\ Accuracy\end{tabular}} & \multirow{2}{*}{\begin{tabular}[c]{@{}c@{}}Adversarial\\ Accuracy\end{tabular}} & \multirow{2}{*}{\begin{tabular}[c]{@{}c@{}}Robustness\\ (w/ patch)\end{tabular}} \\
 & \multicolumn{1}{c|}{} &  &  &  \\ \hline %\hline 
\multirow{3}{*}{\begin{tabular}[c]{@{}c@{}}38\\ x\\ 38\end{tabular}} & ResNet 152 & 94.1\%  & 48.6\%  & 81.7\%  \\ %\cline{2-5} 
 & ResNet 50 & 90.9\%  & 49.2\%  & 76.5\%  \\ %\cline{2-5} 
 & VGG 19 & 88.6\%  & 47.1\%  & 71.9\%  \\ \hline
\multirow{3}{*}{\begin{tabular}[c]{@{}c@{}}41\\ x\\ 41\end{tabular}} & ResNet 152 & 94.1\% & 30.1\% & 78.8\% \\ %\cline{2-5} 
 & ResNet 50 & 90.9\% & 29.3\% & 74.1\% \\ %\cline{2-5} 
 & VGG 19 & 88.6\% & 28.2\% & 75.9\% \\ \hline
\multirow{3}{*}{\begin{tabular}[c]{@{}c@{}}44\\ x\\ 44\end{tabular}} & ResNet 152 & 94.1\%  & 10.8\%  & 80.5\%  \\ %\cline{2-5} 
 & ResNet 50 & 90.9\%  & 11.2\%  & 75.3\%  \\ %\cline{2-5} 
 & VGG 19 & 88.6\%  & 12.6\%  & 75.1\%  \\ \hline
\multirow{3}{*}{\begin{tabular}[c]{@{}c@{}}47\\ x\\ 47\end{tabular}} & ResNet 152 & 94.1\% & 9.1\% & 77.4\% \\ %\cline{2-5} 
 & ResNet 50 & 90.9\% & 9.6\% & 75.4\% \\ %\cline{2-5} 
 & VGG 19 & 88.6\% & 10.4\% & 73.1\% \\ \hline
\multirow{3}{*}{\begin{tabular}[c]{@{}c@{}}50\\ x\\ 50\end{tabular}} & ResNet 152 & 94.1\%  & 6.8\%  & 77.4\%  \\ %\cline{2-5} 
 & ResNet 50 & 90.9\%  & 5.9\%  & 76.1\%  \\ %\cline{2-5} 
 & VGG 19 & 88.6\%  & 6.2\%  & 71.3\%  \\ \hline
\end{tabular}
}
\end{table}

\begin{table}[!htbp]
\caption{Robustness on LAVAN attack (CalTech-101)}
\label{tab2_caltech_sup}
\resizebox{0.45\textwidth}{!}{%
\begin{tabular}{c|l|c|c|c}
\hline
\multirow{2}{*}{\begin{tabular}[c]{@{}c@{}}Patch\\ Size\end{tabular}} & \multicolumn{1}{c|}{\multirow{2}{*}{\begin{tabular}[c]{@{}c@{}}Model /\\ Neural Network\end{tabular}}} & \multirow{2}{*}{\begin{tabular}[c]{@{}c@{}}Baseline\\ Accuracy\end{tabular}} & \multirow{2}{*}{\begin{tabular}[c]{@{}c@{}}Adversarial\\ Accuracy\end{tabular}} & \multirow{2}{*}{\begin{tabular}[c]{@{}c@{}}Robustness\\ (w/ patch)\end{tabular}} \\
 & \multicolumn{1}{c|}{} &  &  &  \\ \hline
\multirow{3}{*}{\begin{tabular}[c]{@{}c@{}}38\\ x\\ 38\end{tabular}} & ResNet 152 & 94.1\%  & 15.6\%  & 77.2\%  \\ %\cline{2-5} 
 & ResNet 50 & 90.9\%  & 17.1\%  & 73.6\%  \\ %\cline{2-5} 
 & VGG 19 & 88.6\%  & 15.3\%  & 73.1\%  \\ \hline
\multirow{3}{*}{\begin{tabular}[c]{@{}c@{}}41\\ x\\ 41\end{tabular}} & ResNet 152 & 94.1\% & 14.2\% & 79.2\% \\ %\cline{2-5} 
 & ResNet 50 & 90.9\% & 13.9\% & 74.2\% \\ %\cline{2-5} 
 & VGG 19 & 88.6\% & 13.8\% & 72.2\% \\ \hline
\multirow{3}{*}{\begin{tabular}[c]{@{}c@{}}44\\ x\\ 44\end{tabular}} & ResNet 152 & 94.1\%  & 8.4\%  & 76.8\%  \\ %\cline{2-5} 
 & ResNet 50 & 90.9\%  & 8.9\%  & 77.7\%  \\ %\cline{2-5} 
 & VGG 19 & 88.6\%  & 9.1\%  & 71.2\%  \\ \hline
\multirow{3}{*}{\begin{tabular}[c]{@{}c@{}}47\\ x\\ 47\end{tabular}} & ResNet 152 & 94.1\% & 5.1\% & 76.8\% \\ %\cline{2-5} 
 & ResNet 50 & 90.9\% & 4.9\% & 73.4\% \\ %\cline{2-5} 
 & VGG 19 & 88.6\% & 6.1\% & 73.8\% \\ \hline
\multirow{3}{*}{\begin{tabular}[c]{@{}c@{}}50\\ x\\ 50\end{tabular}} & ResNet 152 & 94.1\%  & 1.2\%  & 77.5\%  \\ %\cline{2-5} 
 & ResNet 50 & 90.9\%  & 1.0\%  & 74.8\%  \\ %\cline{2-5} 
 & VGG 19 & 88.6\%  & 1.8\%  & 73.9\%  \\ \hline
\end{tabular}
}
\end{table}

\subsection{Comparison with SOTA}
We conduct a comparative analysis of our defense mechanism against LGS \cite{naseer2019local} and Jujutsu \cite{Jujutsu}. Additionally, for comparative purposes, we compare our technique with two certified defenses—namely, De-randomized smoothing (DS) \cite{levine2020randomized} and PatchGuard \cite{xiang2021patchguard}.

As demonstrated in Table \ref{tab:comaprison}, our defense strategy outperforms state-of-the-art techniques. Specifically, our approach achieves a robust accuracy of $67.1\%$, surpassing LGS, Jujutsu, and Jedi defenses, which achieve robust accuracy rates of $53.86\%$, $60\%$, and $64.34\%$, respectively.

\begin{table}[]
    \centering
    \caption{Performance of our proposed defense compared to four state-of-the-art defenses against GoogleAp \cite{googleap} attack.}
    \begin{tabular}{c|c}
    \hline
     Defense   &  Robust Accuracy\\
     \hline
     LGS \cite{naseer2019local}    &  53.86\% \\
     \hline
     DS \cite{levine2020randomized}    & 35.02\% \\
     \hline
     PatchGuard \cite{xiang2021patchguard}  & 30.96\%\\
     \hline
     Jujutsu \cite{Jujutsu}  &  60\%\\
     \hline
     \textbf{Ours}   & \textbf{67.1\%} \\
     \hline
    \end{tabular}

    \label{tab:comaprison}
\end{table}

\subsection{Key Findings}
The key conclusions from the experimental observations are mentioned here:
\begin{itemize}
    \item The proposed defense technique against adversarial patch attacks is successful in providing robustness to the image classification task, with up to $28\%$ recovery for the standard 38 x 38 pixel patch size. 
    \item The performance is consistent across different patch sizes, for various neural architectures and datasets. 
    \item It is also able to outperform the state-of-the-art with $67.1\%$ robust accuracy for ResNet-50 on the ImageNet dataset.
\end{itemize}
%\section{Discussion}
\section{Potential Adaptive Attack}
%\textcolor{red}{PLS CHECK THE CODE IN THE SHARED NOTEBOOK}
In order to generate a patch capable of circumventing our defense, the adversarial noise must adhere to a distribution similar/close to that of clean images.
We implement the adaptive attack by constraining the distribution of the adversarial patch to closely match the average distribution of $n$ randomly selected fragments from the clean image. Specifically, we calculate the average distribution in terms of both mean and standard deviation. We then compute the mean difference and standard deviation ratio between the average distribution of the fragments and the adversarial patch. %Only optimizer steps that result in a patch satisfying the following constraint are preserved. For $0 \leq |\mu-m|\leq 0.5$ and  $0.7 \leq \sigma/std \leq 1.3$ the generated patch was only able to decrease model accuracy from 78\% to 69\% for ResNet50 on ImageNet.

%To implement an adaptive attack, we try to counter the basis of outlier detection. We map the variability of the adversarial patch to the variability of a collection of randomly chosen fragments $x_{i}$, across all three channels. Fitting Gaussian distributions to each of the fragments, we consider an average of the distribution parameters, mean ($m$) and standard deviation ($std$). For generating the patch, during each step taken by the optimizer, we restrict the updates to fit the distribution of the patch (with parameters $\mu$ and $\sigma$) to match the averaged parameters ($m$ and $std$), with some relaxations, with tolerance (dataset specific) of up to $0 \leq |\mu-m|\leq 0.5$  and $1 \leq \sigma/std \leq 2$. Beyond these ranges, the Gaussian distributions are statistically separable. At the extremities, for ResNet50 on ImageNet, the accuracy  drops only slightly from $78\%$ (clean) to $73\%$, and the robust accuracy is $76\%$. 

\begin{figure}[htbp]
\centerline{\includegraphics[width=\columnwidth]{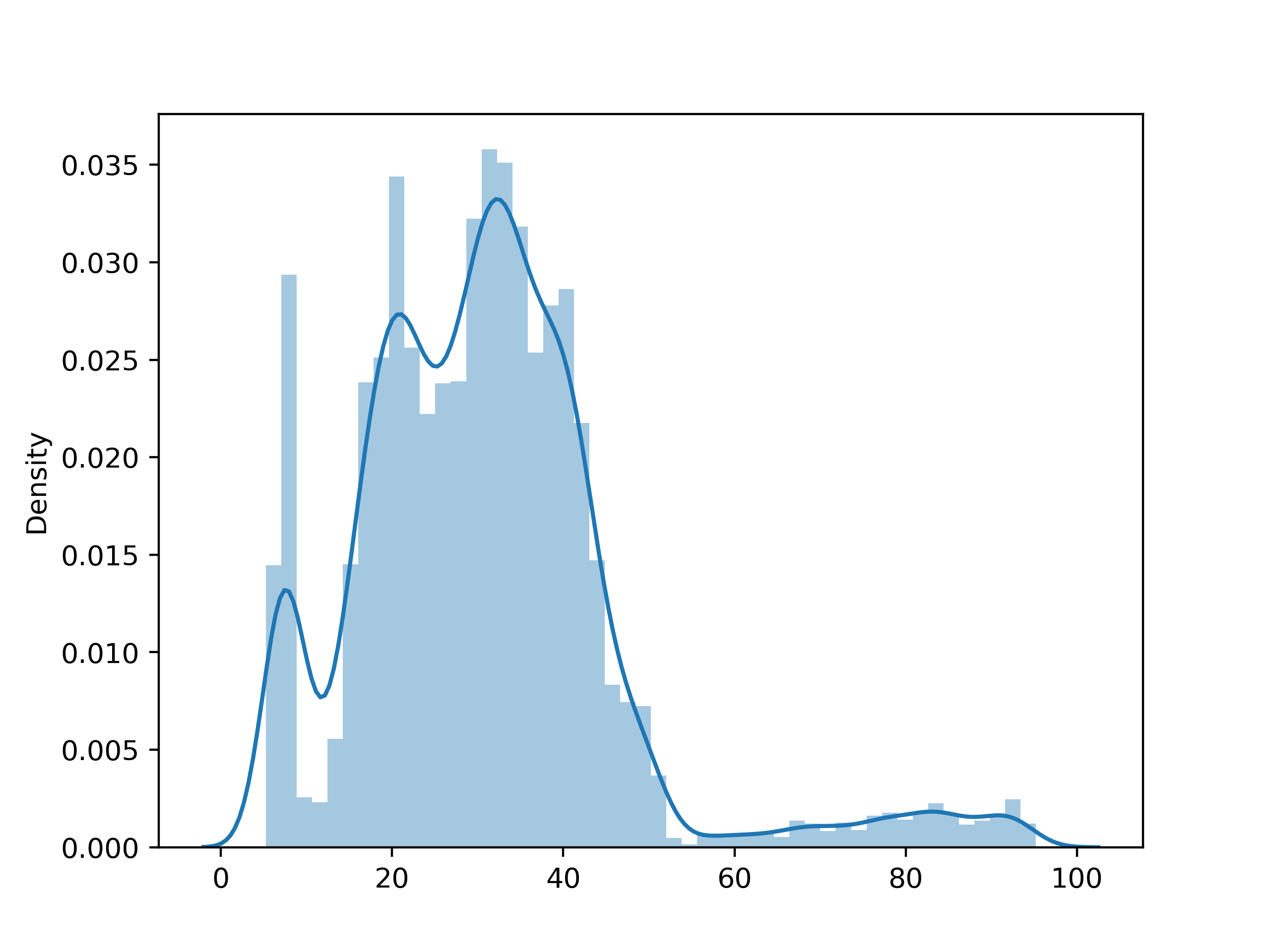}}
\caption{Plot of Mahalanobis distances of segments upon introducing the adaptive attack, showing one single heavy tailed distribution.}
\label{fig:mahal_ada}
\end{figure}

We compute the mean and standard deviation separately for each channel of the patch, and adjust the gradient of each channel accordingly. This ensures that the mean and standard deviation constraints are enforced for each channel of the adversarial patch during optimization.
$0.02\leq|\mu-m|\ \leq 0.08$
$1.5 \leq \sigma/std \leq 2.4$
We recalculated the Mahalanobis distance for each segment and plotted them. Upon observation, we noticed a one single heavy tailed distribution indicating that the segments located on the adversarial patch exhibited a Mahalanobis distance closer to that of the rest of the image segments (See Figure \ref{fig:mahal_ada}) which makes it challenging to isolate the patch as anomalies. %This observation indicates that the patch contains informational variability closer to that of the rest of the image, making it challenging to isolate as anomalies. 
In addition, the generated patch was only able to decrease model accuracy from 78\% to 67\% for ResNet50 on ImageNet, demonstrating that it is difficult to generate an adversarial patch that will not behave as an anomaly as compared to the distribution of the image. 

\section{Related Works}
\label{related}

% Defenses against adversarial patch-based attacks can be categorized into two main approaches: certified defenses and empirical defenses.

% \textit{Certified defenses:} \textbf{De-randomized smoothing (DS)} \cite{levine2020randomized} introduces a certified defense technique by building a smoothed classifier by ensembling local predictions made on pixel patches. \textbf{PatchGuard}\cite{xiang2021patchguard} uses enforcing a small receptive field within deep neural networks (DNNs) and securing feature aggregation by masking out the regions with the highest sum of class evidence. 

% \textit{Empirical defenses:} \textbf{Localized Gradient Smoothing (LGS)} \cite{naseer2019local} first normalizes gradient values and then uses a moving window to identify high-density regions based on certain thresholds.
% \textbf{Jujutsu} \cite{Jujutsu} focuses on localizing adversarial patches and distinguishing them from benign samples. 

% These defenses, while valuable, do come with certain limitations such as high false positive rates, poor detection rates etc.
% Another challenge is the ability to mitigate the adversarial impact effectively while still allowing deep neural networks (DNNs) to make correct inferences on the clean examples. 
Defenses against adversarial patch-based attacks can be broadly classified into two categories: certified defenses and empirical defenses.

\textit{Certified defenses:} \textbf{De-randomized Smoothing (DS)} \cite{levine2020randomized} introduces a certified defense technique by building a smoothed classifier through ensembling local predictions made on pixel patches. \textbf{PatchGuard} \cite{xiang2021patchguard} employs a small receptive field within deep neural networks (DNNs) and secures feature aggregation by masking out regions with the highest sum of class evidence.

\textit{Empirical defenses:} \textbf{Localized Gradient Smoothing (LGS)} \cite{naseer2019local} normalizes gradient values and utilizes a moving window to identify high-density regions based on specific thresholds. \textbf{Jujutsu} \cite{Jujutsu} focuses on localizing adversarial patches and distinguishing them from benign samples.

While these defenses offer valuable contributions, they are not without limitations, including high false positive rates and poor detection rates. Additionally, a significant challenge lies in effectively mitigating the adversarial impact while ensuring deep neural networks (DNNs) make accurate inferences on clean examples.
%Limitations:
%High false positive rates: unable to correctly distinguish adversarial examples from clean samples.
%Poor detection rates: unable to reliably locate the patch.
%Low mitigation rate (robust accuracy):  unable to allow the DNNs to make correct inference on the adversarial examples as many important features are corrupted

\section{Conclusion}
Adversarial patch based attacks are a common and practical mode of exposing vulnerabilities in trained neural networks to behave erroneously for image classification tasks. In this paper, we have proposed a defense mechanism that makes use of the insight that the said adversarial patches contain information or variability which is significantly different from the data distribution in the rest of the image. This helps us in using clustering based anomaly detection techniques to isolate the patches. We have proposed a three-step pipeline for the same and have reported impressive performance. In the future, the objective is to make this technique more robust by developing stronger anomaly detection techniques and extending the defense mechanism to other computer vision tasks.
%==========================================================

\bibliographystyle{IEEEtran}
\bibliography{advnnbib}

\begin{thebibliography}{10}
\providecommand{\url}[1]{#1}
\csname url@samestyle\endcsname
\providecommand{\newblock}{\relax}
\providecommand{\bibinfo}[2]{#2}
\providecommand{\BIBentrySTDinterwordspacing}{\spaceskip=0pt\relax}
\providecommand{\BIBentryALTinterwordstretchfactor}{4}
\providecommand{\BIBentryALTinterwordspacing}{\spaceskip=\fontdimen2\font plus
\BIBentryALTinterwordstretchfactor\fontdimen3\font minus \fontdimen4\font\relax}
\providecommand{\BIBforeignlanguage}[2]{{%
\expandafter\ifx\csname l@#1\endcsname\relax
\typeout{** WARNING: IEEEtran.bst: No hyphenation pattern has been}%
\typeout{** loaded for the language `#1'. Using the pattern for}%
\typeout{** the default language instead.}%
\else
\language=\csname l@#1\endcsname
\fi
#2}}
\providecommand{\BIBdecl}{\relax}
\BIBdecl

\bibitem{naseer2019local}
M.~Naseer, S.~Khan, and F.~Porikli, ``Local gradients smoothing: Defense against localized adversarial attacks,'' in \emph{2019 IEEE Winter Conference on Applications of Computer Vision (WACV)}.\hskip 1em plus 0.5em minus 0.4em\relax IEEE, 2019, pp. 1300--1307.

\bibitem{Jujutsu}
Z.~Chen, P.~Dash, and K.~Pattabiraman, ``Jujutsu: A two-stage defense against adversarial patch attacks on deep neural networks,'' in \emph{Proceedings of the 2023 ACM Asia Conference on Computer and Communications Security}, ser. ASIA CCS '23.\hskip 1em plus 0.5em minus 0.4em\relax New York, NY, USA: Association for Computing Machinery, 2023, p. 689–703.

\bibitem{guesmi2023physical}
A.~Guesmi, M.~A. Hanif, B.~Ouni, and M.~Shafique, ``Physical adversarial attacks for camera-based smart systems: Current trends, categorization, applications, research challenges, and future outlook,'' \emph{IEEE Access}, 2023.

\bibitem{CW}
\BIBentryALTinterwordspacing
N.~Carlini and D.~A. Wagner, ``Towards evaluating the robustness of neural networks,'' \emph{CoRR}, vol. abs/1608.04644, 2016. [Online]. Available: \url{http://arxiv.org/abs/1608.04644}
\BIBentrySTDinterwordspacing

\bibitem{LiV15}
\BIBentryALTinterwordspacing
B.~Li and Y.~Vorobeychik, ``Scalable optimization of randomized operational decisions in adversarial classification settings,'' in \emph{Proceedings of the Eighteenth International Conference on Artificial Intelligence and Statistics, {AISTATS} 2015, San Diego, California, USA, May 9-12, 2015}, ser. {JMLR} Workshop and Conference Proceedings, G.~Lebanon and S.~V.~N. Vishwanathan, Eds., vol.~38.\hskip 1em plus 0.5em minus 0.4em\relax JMLR.org, 2015. [Online]. Available: \url{http://proceedings.mlr.press/v38/li15a.html}
\BIBentrySTDinterwordspacing

\bibitem{Goodfellow2015ExplainingAH}
I.~J. Goodfellow, J.~Shlens, and C.~Szegedy, ``Explaining and harnessing adversarial examples,'' \emph{CoRR}, vol. abs/1412.6572, 2015.

\bibitem{cod_1}
N.~Chattopadhyay, A.~Chattopadhyay, S.~S. Gupta, and M.~Kasper, ``Curse of dimensionality in adversarial examples,'' in \emph{2019 International Joint Conference on Neural Networks (IJCNN)}.\hskip 1em plus 0.5em minus 0.4em\relax IEEE, 2019, pp. 1--8.

\bibitem{cod_2}
N.~Chattopadhyay, S.~Chatterjee, and A.~Chattopadhyay, ``Robustness against adversarial attacks using dimensionality,'' in \emph{International Conference on Security, Privacy, and Applied Cryptography Engineering}.\hskip 1em plus 0.5em minus 0.4em\relax Springer, 2021, pp. 226--241.

\bibitem{guesmi2023dap}
A.~Guesmi, R.~Ding, M.~A. Hanif, I.~Alouani, and M.~Shafique, ``Dap: A dynamic adversarial patch for evading person detectors,'' 2023.

\bibitem{guesmi2023advart}
A.~Guesmi, I.~M. Bilasco, M.~Shafique, and I.~Alouani, ``Advart: Adversarial art for camouflaged object detection attacks,'' \emph{arXiv preprint arXiv:2303.01734}, 2023.

\bibitem{Hu21}
Y.-C.-T. Hu, J.-C. Chen, B.-H. Kung, K.-L. Hua, and D.~S. Tan, ``Naturalistic physical adversarial patch for object detectors,'' in \emph{2021 IEEE/CVF International Conference on Computer Vision (ICCV)}, 2021, pp. 7828--7837.

\bibitem{guesmi2024saam}
A.~Guesmi, M.~A. Hanif, B.~Ouni, and M.~Shafique, ``Saam: Stealthy adversarial attack on monocular depth estimation,'' \emph{IEEE Access}, 2024.

\bibitem{guesmi2023aparate}
A.~Guesmi, M.~A. Hanif, I.~Alouani, and M.~Shafique, ``Aparate: Adaptive adversarial patch for cnn-based monocular depth estimation for autonomous navigation,'' \emph{arXiv preprint arXiv:2303.01351}, 2023.

\bibitem{xiang2021patchguard}
C.~Xiang, A.~N. Bhagoji, V.~Sehwag, and P.~Mittal, ``$\{$PatchGuard$\}$: A provably robust defense against adversarial patches via small receptive fields and masking,'' in \emph{30th USENIX Security Symposium (USENIX Security 21)}, 2021, pp. 2237--2254.

\bibitem{levine2020randomized}
A.~Levine and S.~Feizi, ``(de) randomized smoothing for certifiable defense against patch attacks,'' \emph{Advances in Neural Information Processing Systems}, vol.~33, pp. 6465--6475, 2020.

\bibitem{dbscan}
M.~Ester, H.-P. Kriegel, J.~Sander, X.~Xu \emph{et~al.}, ``A density-based algorithm for discovering clusters in large spatial databases with noise,'' in \emph{kdd}, vol.~96, no.~34, 1996, pp. 226--231.

\bibitem{dbscan_2}
E.~Schubert, J.~Sander, M.~Ester, H.~P. Kriegel, and X.~Xu, ``Dbscan revisited, revisited: why and how you should (still) use dbscan,'' \emph{ACM Transactions on Database Systems (TODS)}, vol.~42, no.~3, pp. 1--21, 2017.

\bibitem{lavan}
D.~K. et~al., ``Lavan: Localized and visible adversarial noise,'' in \emph{International Conference on Machine Learning}, 2018.

\bibitem{googleap}
\BIBentryALTinterwordspacing
T.~B. et~al., ``Adversarial patch,'' 2017. [Online]. Available: \url{https://arxiv.org/pdf/1712.09665.pdf}
\BIBentrySTDinterwordspacing

\bibitem{eot}
A.~Athalye, L.~Engstrom, A.~Ilyas, and K.~Kwok, ``Synthesizing robust adversarial examples,'' in \emph{International conference on machine learning}.\hskip 1em plus 0.5em minus 0.4em\relax PMLR, 2018, pp. 284--293.

\bibitem{mahal_1}
K.~I. Penny, ``Appropriate critical values when testing for a single multivariate outlier by using the mahalanobis distance,'' \emph{Journal of the Royal Statistical Society: Series C (Applied Statistics)}, vol.~45, no.~1, pp. 73--81, 1996.

\bibitem{mahal}
R.~De~Maesschalck, D.~Jouan-Rimbaud, and D.~L. Massart, ``The mahalanobis distance,'' \emph{Chemometrics and intelligent laboratory systems}, vol.~50, no.~1, pp. 1--18, 2000.

\bibitem{mahal_2}
G.~J. McLachlan, ``Mahalanobis distance,'' \emph{Resonance}, vol.~4, no.~6, pp. 20--26, 1999.

\bibitem{dbscan_3}
D.~Birant and A.~Kut, ``St-dbscan: An algorithm for clustering spatial--temporal data,'' \emph{Data \& knowledge engineering}, vol.~60, no.~1, pp. 208--221, 2007.

\bibitem{dbscan_4}
A.~Smiti and Z.~Elouedi, ``Dbscan-gm: An improved clustering method based on gaussian means and dbscan techniques,'' in \emph{2012 IEEE 16th international conference on intelligent engineering systems (INES)}.\hskip 1em plus 0.5em minus 0.4em\relax IEEE, 2012, pp. 573--578.

\bibitem{imagenet}
J.~Deng, W.~Dong, R.~Socher, L.-J. Li, K.~Li, and L.~Fei-Fei, ``Imagenet: A large-scale hierarchical image database,'' in \emph{2009 IEEE Conference on Computer Vision and Pattern Recognition}, 2009, pp. 248--255.

\bibitem{he2015deep}
K.~He, X.~Zhang, S.~Ren, and J.~Sun, ``Deep residual learning for image recognition,'' 2015.

\bibitem{simonyan2015deep}
K.~Simonyan and A.~Zisserman, ``Very deep convolutional networks for large-scale image recognition,'' 2015.

\end{thebibliography}
\balance
\end{document}